\documentclass{article} % For LaTeX2e
\usepackage{iclr2018_conference,times}
\usepackage{hyperref}
\usepackage{url}

\usepackage{graphicx}
\usepackage{booktabs}
\usepackage{multirow}
\usepackage{diagbox}
\usepackage{amsmath}
\usepackage{xfrac}
\usepackage{subfigure}
\title{Attacking Binarized Neural Networks}

\author{Angus Galloway\textsuperscript{1},~Graham W. Taylor\textsuperscript{1,2,3}~Medhat Moussa\textsuperscript{1} \\
\textsuperscript{1}School of Engineering, University of Guelph, Canada\\
\textsuperscript{2}Canadian Institute for Advanced Research\\
\textsuperscript{3}Vector Institute for Artificial Intelligence, Canada\\
\texttt{\{gallowaa,gwtaylor,mmoussa\}@uoguelph.ca} \\
}

\iclrfinalcopy % Uncomment for camera-ready version, but NOT for submission.
\newif\iffinal
\finaltrue % Uncomment for camera-ready version, but NOT for submission.

\begin{document}

\maketitle

\begin{abstract}
Neural networks with low-precision weights and activations offer compelling
efficiency advantages over their full-precision equivalents. The two most
frequently discussed benefits of quantization are reduced memory consumption,
and a faster forward pass when implemented with efficient bitwise
operations. We propose a third benefit of very low-precision neural networks:
improved robustness against some adversarial attacks, and in the worst case,
performance that is on par with full-precision models. We focus on the very
low-precision case where weights and activations are both quantized to $\pm$1,
and note that stochastically quantizing weights in just one layer can sharply
reduce the impact of iterative attacks. We observe that non-scaled binary neural
networks exhibit a similar effect to the original \emph{defensive distillation}
procedure that led to \emph{gradient masking}, and a false notion of security.
We address this by conducting both black-box and white-box experiments with
binary models that do not artificially mask gradients.\footnote{Source 
code available at \url{https://github.com/AngusG/cleverhans-attacking-bnns}}

\end{abstract}

\section{Introduction}

The ability to fool machine learning models by making small changes
to their input severely limits their potential for safe use in many
real-world scenarios. Example vulnerabilities include a seemingly innocuous
audio broadcast that is interpreted by a speech recognition model in a
smartphone, with the intent to trigger an e-transfer, as well as pictures or identity
documents that are automatically tagged as someone other than the real
individual.

% Was trying to be reasonably creative with above examples and emphasizing that
% adversarial examples are not limited to domain of images, but open to
% feedback on how to make more compelling. You always hear the same autonomous
% driving stop/yield sign example all the time. In reality, autonomous cars
% will not depend on visual cues for the locations of road signs, they will
% be geotagged in maps or use infrastructure to vehicle radios e.g.~traffic
% lights.

The two most common threat models when evaluating the security of a system are
the \emph{black-box} and \emph{white-box} assumptions, which represent
varying degrees of information that an adversary may possess. In a \emph{black-box}
threat model, an adversary has similar abilities to a normal user that
interacts with a system by providing inputs and observing the corresponding
outputs. Under this threat model, an adversary generally does not know
details of the model architecture or dataset used to train the model. Of
course, an adversary is free to assume that a convolutional architecture
was likely used if the input domain is images, or a recurrent model for speech
or text.

In a \emph{white-box} threat model, an adversary has complete access to the
model architecture and parameters. In the case of neural networks, white-box
attacks frequently rely on gradient information to craft especially strong
\emph{adversarial examples}, where \emph{strong} means that the example is very
close to the original input as defined by some distance norm (e.g.~$L_0$--number
of features modified, $L_2$--mean squared distance), yet is very likely
to cause the model to yield the incorrect output. For both threat types,
targeted attacks where a model is made to fail in a specific way (e.g.~causing a
handwritten `7' look like a `3') represents a stronger attack than simple
misclassification.

The problem with deploying machine learning systems that are \emph{secured} in
a traditional sense, is that adversarial examples have been shown to generalize
well between models with different source and target
architectures~\citep{szegedy2013intriguing, Papernot17, Tramer17}.
This means that a secured model can be compromised in an approximately
white-box setting by training and attacking a substitute model that
approximates the decision boundary of the model under
attack~\citep{Papernot17}. Thus, to make strong conclusions about the
robustness of a machine learning model to adversarial attacks, both threat
models should be considered.

Tangent to research on defences against adversarial attacks, significant
progress has been made towards training very low-precision deep neural networks
to accuracy levels that are competitive with full-precision
models~\citep{CourbariauxB16, zhou2016dorefa, Tang_train_bnn_17}. The current motivation for
extreme quantization is the ability to deploy these models under
hardware resource constraints, acceleration, or reduced power consumption. 
Ideally, 32$\times$ compression is possible by using 1-bit to represent 
single-precision floating point parameters. By similarly quantizing
activations, we can reduce run-time memory consumption as well. These savings
enable large scale deployment of neural networks on the billions of existing
embedded devices. Very low-precision models were designed with
deployment in mind, and may be responsible for making critical decisions in
embedded systems, all subject to reverse engineering and a diverse set of
real world attacks. With much at stake in applications like autonomous
navigation, robotics, and network infrastructure, understanding how very
low-precision neural networks behave in adversarial settings is essential. To
that end, we make the following contributions:

\begin{itemize}

	\item To the best of our knowledge, we are the first to formally evaluate
	and interpret the robustness of binary neural networks (BNNs) to adversarial
	attacks on the MNIST~\citep{mnist98} and
	CIFAR-10~\citep{Krizhevsky09learningmultiple} datasets.

	\item We compare and contrast the properties of low-precision neural
	networks that confer adversarial robustness to previously proposed defense
	strategies. We then combine these properties to propose an optimal defense
	strategy.

	\item We attempt to generalize and make recommendations regarding the
	suitability of low-precision neural networks against various classes of
  attacks (e.g.~single step vs.~iterative).

\end{itemize}

\section{Background}

Since the initial disclosure of adversarial examples
by~\citet{szegedy2013intriguing} and~\citet{Biggio2013}, many defense
strategies have been proposed and subsequently defeated. It is
generally accepted that strategies for mitigating the impact of these examples
still lag behind
state of the art attacks, which are capable of producing adversarial examples that are
indistinguishable from unmodified
inputs as perceived by humans. In general, there are two approaches to defending
against adversarial examples: \emph{reactive}--detecting the presence of
adversarial examples, such as through some notion of confidence-based outlier
detection. On the other hand, a \emph{proactive} approach
aims to improve the robustness of the underlying model, which may involve
adding an extra class to which malicious inputs should be
assigned~\citep{PapernotM17}. The latter approach is
important for building reliable systems where a sensible
decision~\emph{must} be made at all times. In this work, we focus solely on the
proactive approach.

To define adversarial examples, we require some measurement of distance that
can be computed between perturbed inputs and naturally occurring inputs. In the
visual domain, it is convenient if the metric approximates human perceptual
similarity, but is not required. Various $L_p$ norms have been used in the
literature: $L_0$--number of features modified, $L_2$--mean squared distance,
$L_\infty$--limited only in the maximum perturbation applied to any feature.
We evaluate at least one attack that is cast in terms of each respective
distance metric, and leave discussion of the optimal distance metric to future work.

The most compelling explanation for the existence of adversarial examples
proposed to date is the linearity
hypothesis~\citep{Goodfellow_explaining_adversarial}.
The elementary operators, matrix dot-products and convolutions used at each
layer of a neural network are fundamentally too linear. Furthermore, the
non-linearity applied at each layer is usually itself either piecewise linear
(e.g.~ReLU), or we have specifically encouraged the network through
initialization
or regularization to have small weights and
activations such that its units (e.g.~sigmoid, tanh) operate in their linear regions. By adding
noise to inputs which is highly correlated with the sign of the model
parameters, a large swing in activation can be induced. Additionally, the
magnitude by which this noise must be scaled to have this effect tends to diminish
as the input dimensionality grows. This piecewise linearity also makes neural
networks easy to attack using the gradient of the output with respect to the
input, and consequently, the resulting incorrect predictions are made with
high-confidence.

Fortunately, we are reminded that the universal approximation theorem suggests
that given sufficient capacity, a neural network should at least be able to
represent the type of function that resists adversarial
examples~\citep{Goodfellow_explaining_adversarial}. The most successful defense
mechanism to date,
adversarial training, is based on this premise, and attempts to learn such a
function. The \emph{fast gradient sign method} (FGSM) is one such procedure for
crafting this damaging noise, and is still used today despite not being
state-of-the-art in the white-box setting, as it is straightforward to compute,
and yields examples that transfer well between
models~\citep{Goodfellow_explaining_adversarial}.

The linearity hypothesis was one of the main reasons for initially considering
binarized neural networks as a natural defense against adversarial examples.
Not only are they highly regularized by default through severely quantized
weights, but they appear to be more non-linear and discontinuous than
conventional deep neural networks (DNNs). Additionally, we
suspect that the same characteristics making them challenging to train, make
them difficult to attack with an iterative procedure. At the same time,
assumptions regarding the information required by an effective adversary have
become more and more relaxed, to the extent that black-box attacks
can be especially damaging with just a small set of labeled input-output
pairs~\citep{Papernot17}.

Perhaps the most striking feature of adversarial examples is how well they
generalize between models with different architectures while trained on different
datasets~\citep{Goodfellow_explaining_adversarial, Papernot17,Kurakin17_adv_at_scale}. It was shown
by~\citet{Kurakin17_physical_world} that 2/3 of adversarial
ImageNet examples survive various camera and perspective transformations after being
printed on paper and subsequently photographed and classified by a mobile phone.

The most successful black-box attacks have the secured model (Oracle) assign
labels to a set of real or synthetic inputs, which can be used to train a
substitute model that mimics the Oracle's decision boundary~\citep{Papernot17}.
A single step attack, such as FGSM, can be used on the smooth substitute model to
generate examples that transfer, without having access to the original training
data, architecture, or training procedure used
by the Oracle.~\citet{Papernot17} showed they are able to compromise machine
learning models 80\% of the time on small datasets like MNIST using
various shallow MLP-based substitute models. There is not a particularly high
correlation between test accuracy and transferability of adversarial examples;
therefore despite not attaining great results on the original MNIST task, a
simple substitute learns enough to compromise the Oracle. This technique was
shown to overcome gradient masking approaches, such as in the case with models
that either obscure or have no gradient information, such as k-nearest neighbors
or decision trees.

With strong adversarial training of the model to be defended, attacks generated
using the substitute model do not transfer as well. Therefore, to be compelling,
BNNs should be able to handle training with large $\epsilon$ while
maintaining competitive test accuracy on clean inputs relative to full-precision.

The strongest white-box attacks all use an iterative procedure; however, the
resulting examples do not transfer as well as single step
methods~\citep{Goodfellow_explaining_adversarial}.
An iterative attack using the Adam optimizer was proposed
by~\citet{carlini2017towards} that outperforms other expensive optimization
based approaches~\citet{szegedy2013intriguing}, the Jacobian-based saliency
map attack (JSMA)~\citep{PapernotMJFCS15}, and Deepfool~\citep{DeepFool} in terms
of three $L_p$ norms previously used as an adversarial example distance metrics
in the literature. We have made our best attempt to use state-of-the-art attacks
in our experiments.

\section{Experiments}

\begin{figure}[h]
\begin{center}
\includegraphics[width=\linewidth]{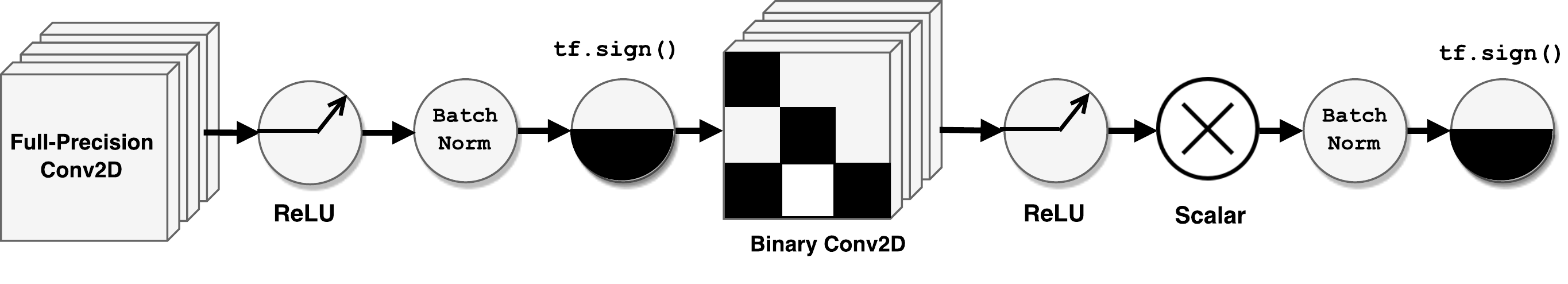}
\end{center}
\caption{Blocks used in binary convolution architecture.}
\label{fig:bin_layer}
\end{figure}

In Figure~\ref{fig:bin_layer}, we depict the quantization scheme applied to
the base convolutional neural network provided in the CleverHans library
tutorials~\citep{papernot2017cleverhans}. 
In the first layer, we retain weights and activations in single-precision floating point.
Weights in hidden layers are binarized either deterministically or stochastically, as
in~\citet{CourbariauxB16}, and activations were always binarized deterministically.
Unlike in~\citet{CourbariauxB16}, we stochastically quantize weights at
\emph{test} time as a possible defense against iterative attacks. Under the
stochastic binarization scheme, weights are sampled once per forward pass from
a Bernoulli distribution with probability given by passing the real valued
weight through the hard sigmoid function from~\citet{CourbariauxB16}. Lastly, we
map the Bernoulli samples $\in [0,1]$ to $\pm1$ by multiplying by 2 and
subtracting 1\footnote{In TensorFlow this can be accomplished with:\\
\texttt{2 * Bernoulli(probs=tf.clip\_by\_value((x + 1.)/ 2., 0., 1.))).sample() % chktex 9 % chktex 36 
-1}}.% chktex 10
We do not find that this significantly slows down training with
TensorFlow~\citep{tensorflow2015-whitepaper} on a modern GPU, but these
networks take between 3--4$\times$ as many epochs as a deterministically quantized binary
network to converge.

We use the straight through estimator (STE) to back-propagate gradients through
the quantization step~\citep{BengioLC13}. We optionally insert a small (e.g 1e-2) tunable
scalar after the ReLU in hidden layers, to compensate for an increase in the L1 norm of
the activations due to binarization.~\citet{Tang_train_bnn_17} also used this approach 
to reach similar accuracy gains as those conferred by the more expensive XNOR-Net
channel-wise normalization scheme~\citep{rastegariECCV16}.
Convolution kernels were initialized from a truncated normal distribution with
$\sigma$=0.2 for accumulating full-precision weight updates, and were quantized
to $\pm1$ in the forward pass. Batch normalization was applied before
quantizing activations to ensure they were centered around zero~\citep{batch_norm}.

We report test error rates for these models on MNIST~\citep{mnist98} with
varying capacity in Table~\ref{table:models-and-clean-acc} of
Appendix~\ref{sec:clean-test-error}. Capacity is denoted by the number of
kernels in the first layer, $K_{Layer1}$. All subsequent layers had exactly
double this number of kernels. Models were trained for 15 epochs unless
indicated otherwise. In general, models with full-precision weights and
activations under-fit the naturally occurring data less than a binary
equivalent, with error rates of approximately 1\% and 2\%, respectively.
With the addition of the small learned scaling factor, the binary models converge
to approximately the same error rate as the full-precision model on MNIST
and CIFAR-10\@.

We experiment with three different types of adversarial training, depending on
the combination of dataset and attack: FGSM with fixed $\epsilon$, FGSM with
$\epsilon$ sampled from a truncated normal
distribution as in~\citet{Kurakin17_adv_at_scale}, and projected gradient
descent (PGD)~\citep{madryetal}, which is the state-of-the-art adversarial
training procedure for MNIST\@. We do not necessarily pair all training methods
against all attacks. The model's own best prediction is used as the true label
to minimize in adversarial training, unless otherwise noted to prevent the
\emph{label leaking} effect~\citep{Kurakin17_adv_at_scale}. We first attempt to
fool our binarized networks with single step attacks in a white-box setting,
and progressively scale up to stronger state-of-the-art attacks. All 
experiments were conducted by seeding the TensorFlow random number generator 
with the value 1234.

\subsection{White-Box Attacks}

All experiments were conducted in TensorFlow, and used either v2.0.0 of
CleverHans~\citep{papernot2017cleverhans}, or Foolbox, a Python toolbox for
creating adversarial examples~\citep{rauber2017foolbox}. All attacks were
clipped to the anticipated input range during adversarial training and evaluation.
For single step attacks, we fix the magnitude of the perturbation 
and attack the whole test set, then report accuracy on the new test set.
The general procedure for iterative attacks is to fix the step size per iteration
or learning rate, and number of iterations. We then report accuracy on the perturbed
test set after this many iterations while keeping other hyper-parameters constant.

\subsubsection{Fast Gradient Sign Method}

The FGSM is a simple but effective single step
attack first introduced in~\citet{Goodfellow_explaining_adversarial}, and
defined in eq~\eqref{eq:fgsm}. The attack linearly approximates the gradient of
the loss used to train the model with respect to the input. The gradient is
thresholded by taking its sign, scaled by a uniform constant $\epsilon$, and
added to, or subtracted from, the input, depending on if we wish to minimize the
current class, or move in the direction of a target class:

\begin{equation} \label{eq:fgsm}
	x_{adv} = x + \epsilon \times sign (\Delta_x J(\theta, x, y))
\end{equation}

To confer robustness to more than one value of $\epsilon$ with which an
adversary may use to attack, the adversarial training procedure
from~\citet{Kurakin17_adv_at_scale} proposes to sample a unique $\epsilon$ for
each training example from a truncated normal distribution. We set the standard
deviation to $\sigma$ = $ceil(\epsilon_{\max} * 255 / 2)$. We consider up to
$\epsilon_{\max}=0.3$, as this is a common upper limit for a $L_\infty$ norm
perturbation that is not easily perceived by humans, and corresponds to a 30\%
change in pixel intensity for an arbitrary number of pixels.

\begin{table}[h!]
\centering
\caption{Accuracy on adversarial examples generated with a FGSM misclassification
attack on the MNIST test set with three values of $\epsilon$. Three different
models were evaluated: A is full-precision, B is binary, and C is binary with a
learned scalar.
Models trained with, and without, adversarial training are shown. The `+' suffix
indicates the model was trained for the last 5 epochs with the procedure
from~\citet{Kurakin17_adv_at_scale}. All values averaged over four runs for
models trained from scratch.}
\label{table:fgsm-sweep-eps}
\begin{tabular}{|c|c|c|c|c|}
\hline
Model                 & $K_{Layer 1}$ & $\epsilon=0.1$ &
$\epsilon=0.2$ & $\epsilon=0.3$
\\ \hline\hline
\multirow{3}{*}{A}    & 64      & 74$\pm$4\%     & 39$\pm$4\%     & 22$\pm$5\%
\\ \cline{2-5}
                      & 128     & 75$\pm$3\%     & 34$\pm$2\%     & 18$\pm$3\%
                      \\ \cline{2-5}
                      & 256     & 74$\pm$1\%     & 33$\pm$2\%     & 17$\pm$3\%
                      \\ \hline
\multirow{3}{*}{B}    & 64      & 75$\pm$2\%     & 64$\pm$3\%     & 59$\pm$2\%
\\ \cline{2-5}
                      & 128     & 85$\pm$1\%     & 77$\pm$2\% & 70$\pm$2\%
                      \\ \cline{2-5}
                      & 256     & \textbf{89$\pm$1\%} & \textbf{83$\pm$1\%}
                      &
                      \textbf{78$\pm$1\%}
                      \\ \hline
\multirow{3}{*}{C}  & 64      & 56$\pm$7\%     & 27$\pm$5\%     & 15$\pm$3\%
\\ \cline{2-5}
                      & 128     & 64$\pm$3\%     & 26$\pm$9\%     & 11$\pm$5\%
                      \\ \cline{2-5}
                      & 256     & 73$\pm$2\%     & 37$\pm$6\%     & 16$\pm$3\%
                      \\ \hline\hline
\multirow{3}{*}{A+}   & 64      & 80$\pm$1\% & 62$\pm$1\%     & 63$\pm$1\%
\\ \cline{2-5}
                      & 128     & \textbf{83$\pm$1\%}     &
                      \textbf{71$\pm$1\%}     &
                      \textbf{72$\pm$1\%}
                      \\ \cline{2-5}
                      & 256     & \textbf{83$\pm$1\%}     &
                      \textbf{71$\pm$2\%}     &
                      70$\pm$2\%
                      \\ \hline
\multirow{3}{*}{B+}   & 64      & 68$\pm$1\%     & 32$\pm$5\%     & 31$\pm$5\%
\\ \cline{2-5}
                      & 128     & 75$\pm$1\%     & 50$\pm$3\%     & 45$\pm$4\%
                      \\ \cline{2-5}
                      & 256     & 79$\pm$2\%     & 64$\pm$3\%     & 58$\pm$2\%
                      \\ \hline
\multirow{3}{*}{C+} & 64      & 80$\pm$2\%     & 47$\pm$7\%     & 38$\pm$4\%
\\ \cline{2-5}
                      & 128     & 82$\pm$1\%     & 50$\pm$3\%     & 40$\pm$2\%
                      \\ \cline{2-5}
                      & 256     & \textbf{84$\pm$3\%}     & 54$\pm$4\%     &
                      41$\pm$4\%
                      \\ \hline
\end{tabular}
\end{table}

%\vspace{-2mm}

In Table~\ref{table:fgsm-sweep-eps}, it can be observed that a plain binary
network without adversarial training (B) achieves the best robustness to FGSM,
with nearly 90\% accuracy for $\epsilon=0.1$ for the highest capacity model. We
postpone a formal explanation of this outlier for the discussion.
Our results for large $\epsilon$ agree with observations made
by~\citet{madryetal} where they found FGSM to be suboptimal for training as it
yields a limited set of adversarial examples.
We suspect that the reason neither scaled nor unscaled binary models performed
well when trained with an adversary and tested on larger values of $\epsilon$ is
because by the time adversarial training was introduced at epoch 10, both had
entered into a state of decreased learning. Our binary weight implementation
makes updates to real valued weights during training, which are
binarized in the forward pass. The real valued weights tend to
polarize as the model converges, resulting in fewer sign
changes. Regularization schemes actually encouraging the underlying real valued
weights to polarize around $\pm1$ have been
proposed~\citep{Tang_train_bnn_17}, but we do not find this to be particularly
helpful after sweeping a range of settings for the regularization constant
$\lambda$. Regardless, in this case, the binary models did not benefit from
adversarial training to the same extent that the full-precision models did.

We find that adversarial training with binary models is somewhat of a
balancing act. If a strong adversary is introduced to the model too early, it
may fail to converge for natural inputs. If introduced too late, it may be
difficult to bring the model back into its \emph{malleable} state, where it is
willing to flip the sign of its weights. Despite this challenge, the \emph
{scaled} binary model (C+) (see Figure~\ref{fig:bin_layer} for location of 
optional scalar) reaped significant benefits from adversarial
training and its accuracy was on par with the full-precision model for
$\epsilon=0.1$.

To investigate the low performance observed against large $\epsilon$ in
Table~\ref{table:fgsm-sweep-eps}, models A and C were trained from scratch with
40 iterations of PGD~\citep{madryetal}. Table~\ref{table:fgsm-pgd-adv} shows the result of this new training and
subsequent FGSM attack performed identically to that of
Table~\ref{table:fgsm-sweep-eps}.
A similar trend was found in Tables~\ref{table:fgsm-sweep-eps} and~\ref{table:fgsm-pgd-adv}, where the lowest capacity models struggle to become robust
against large $\epsilon$. Once the scaled
binary model had sufficient capacity, it actually slightly outperforms its
full-precision equivalent for all values of $\epsilon$. With this, we have demonstrated that
not only can BNNs achieve competitive accuracy on clean inputs
with significantly fewer resources, but they can also allocate excess capacity in
response to state-of-the-art adversaries.

% \usepackage{multirow}
\begin{table}[h!]
\centering
\caption{Accuracy on adversarial examples generated with a FGSM misclassification
	attack on the MNIST test set with three values of $\epsilon$. Both
	full-precision (A+*) and scaled binary (C+*) models were trained with 40
	iterations of PGD~\citep{madryetal} for the last 5 epochs with with
	$\epsilon=0.3$. All values averaged over four runs for
	models trained from scratch.}
\label{table:fgsm-pgd-adv}
\begin{tabular}{|c|l|c|c|c|}
	\hline
	\multicolumn{1}{|l|}{Model} & $K_{Layer 1}$ & $\epsilon=0.1$ & $\epsilon=0.2$ &
	$\epsilon=0.3$          \\ \hline
	\multirow{3}{*}{A+*}         & 64      & 94.7$\pm$0.2\% & 90.9$\pm$0.3\% &
	80.2$\pm$0.2\% \\ \cline{2-5}
	& 128     & 95.8$\pm$0.3\% & 92.3$\pm$0.3\% &
	82.9$\pm$0.9\% \\ \cline{2-5}
	& 256     & 95.9$\pm$0.2\% & 92.9$\pm$0.3\% &
	85$\pm$1\%     \\ \hline
	\multirow{3}{*}{C+*}         & 64      & 92.9$\pm$0.4\% & 83.6$\pm$0.6\% &
	67$\pm$2\%     \\ \cline{2-5}
	& 128     & 95.0$\pm$0.2\% & 88.2$\pm$0.3\% &
	74.3$\pm$0.6\% \\ \cline{2-5}
	& 256     & \textbf{96.8$\pm$0.3\%} &
	\textbf{93.4$\pm$0.3\%} &
	\textbf{85.6$\pm$0.6\%} \\ \hline
\end{tabular}
\end{table}

\subsubsection{Carlini-Wagner Attack~\cite{carlini2017towards}}
\label{sec:cw}

\begin{figure}[h]
\begin{center}
\includegraphics[width=0.75\linewidth]
{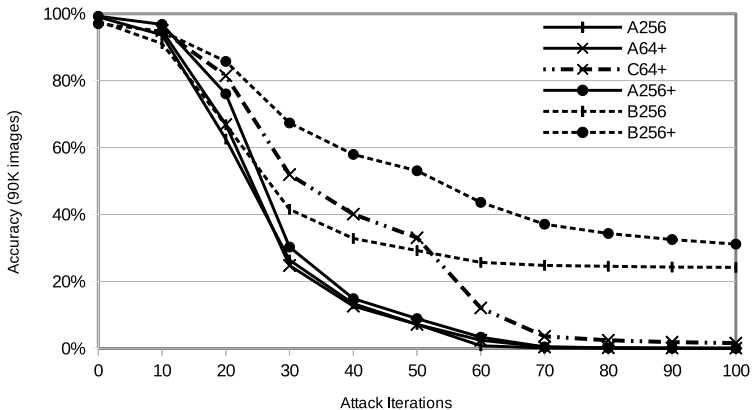}
\end{center}
\caption{Accuracy of full-precision (A), binary (B), and scaled binary (C),
models subject to targeted Carlini-Wagner $L_2$ attacks of increasing strength
on MNIST dataset. Models A/B256+ and A/C64+ were trained with 20 and 40
iterations of PGD, respectively.}
\label{fig:cw-acc-vs-attk-itr}
\end{figure}

The Carlini-Wagner L2 attack~\citet{carlini2017towards} (CWL2) is an 
iterative process guided by an optimizer such as Adam, that produces strong
adversarial examples by simultaneously minimizing distortion, and 
manipulating the logits per the attack goal.
We use the implementation from CleverHans~\citep{papernot2017cleverhans} and show 
results in Table~\ref{table:cw-binary-sweep-capacity} and
Figure~\ref{fig:cw-acc-vs-attk-itr}. Only binary models are shown in
Table~\ref{table:cw-binary-sweep-capacity} because all but two 
full-precision models had \emph{zero} accuracy after running CWL2 for 
100 iterations. The best full-precision model was A256+ with 1.8$\pm$0.9\% accuracy.
We note that the stochastically quantized binary models with scaling to prevent
gradient masking (`S' prefix) underfit somewhat on the training set, and had
test error rates of 8$\pm$1\%, 5$\pm$2\%, and 3$\pm$1\% for each of
S64, S128, and S256 averaged over four runs. For S256, this test error can be 
compared with an unscaled binary model which only achieves 22$\pm$3\%
accuracy \emph{with} gradient masking compared to 46$\pm$3\%~\emph{without}.

%\vspace{-5mm}

\begin{table}[h!]
\centering
\caption{Carlini-Wagner $L_2$ targeted attack on MNIST test
	set (90k images total) for binary models versus increasing capacity from left to
	right. All attacks were run for 100 iterations as all full-precision models
	were driven to have zero accuracy by this point.
	Models with `S' prefix used stochastic quantization.}
\begin{tabular}{||c c c c c||}
\hline
\rule{0pt}{2ex}
Model & B32 & B64 & B128 & B256\\ [0.5ex]
\hline\hline
\rule{0pt}{2ex}
Accuracy & \textbf{7$\pm$1\%} & 7$\pm$3\% & 12$\pm$3\% & 22$\pm$3\% \\
Mean $L_2$ dist. & 2.88$\pm$0.02 & 3.1$\pm$0.2 & 3.2$\pm$0.1 &
3.2$\pm$0.1\\[0.1ex]

\hline\hline
\rule{0pt}{2ex}
Model & B32+ & B64+ & B128+ & B256+\\ [0.5ex]
\hline\hline
\rule{0pt}{2ex}
Accuracy & 3$\pm$1\% & 2.9$\pm$0.6\% & 15$\pm$2\% & 29$\pm$3\%\\
Mean $L_2$ dist. & 3.36$\pm$0.03 & 3.43$\pm$0.05 & 2.9$\pm$0.1 &
2.4$\pm$0.2\\

\hline\hline
\rule{0pt}{2ex}
Model & -- & S64 & S128 & S256\\ % chktex 8
\hline\hline
\rule{0pt}{2ex}
Accuracy & -- & \textbf{71$\pm$2\%} & \textbf{57$\pm$5\%} & % chktex 8
\textbf{46$\pm$3\%}\\
Mean $L_2$ dist. & -- & 1.9$\pm$0.3 & 3.0$\pm$0.4 & 3.5$\pm$0.1 % chktex 8
\\[0.1ex]
\hline
\end{tabular}
\label{table:cw-binary-sweep-capacity}
\end{table}

% I could produce error bars for this plot if it's helpful.

In Figure~\ref{fig:cw-acc-vs-attk-itr}, it can be observed that binary and
full-precision models perform somewhat similarly for the first few iterations
of the CWL2 attack, but beyond 10--20 iterations, the accuracy of
full-precision models
drops off quickly, regardless of having performed adversarial training.
We note that PGD, defined with respect to the $L_\infty$ norm, makes no claim
of increasing robustness to $L_2$ attacks, such as CWL2. Interestingly, it can
be seen that the binary model benefited from adversarial training considerably
when evaluated at 10 to 100 attack iterations, while the full-precision model
did not. These benefits eventually disappear to within the margin of random
error after continuing to 1000 iterations, as recommended by
\cite{carlini2017towards}. At this point, both B and B+ had accuracy of
19$\pm$3\%, by which time the full-precision models had long flatlined at zero.
Meanwhile, S64 maintained $38\pm3$\% accuracy after 1000 iterations, nearly
double that of the deterministically quantized models. Running these attacks to
1000 iterations was two orders of magnitude more time consuming than training
these models from scratch (without PGD training); therefore we believe this
targeted attack represents a fairly substantial level of effort on behalf of
the adversary.

\subsection{Black-Box Attacks}

We run the substitute model training procedure from~\citet{Papernot17}
using CleverHans v2.0.0, for both MNIST and CIFAR-10 datasets with and without
FGSM adversarial training. As a substitute model, we use a two-layer MLP with
200 hidden units and ReLU activations. The substitute is trained on 150 images
withheld from the test set, and augmented by perturbing the images in the direction
of maximal variability of the substitute model, as defined by the Jacobian. Six
epochs of data augmentation with $\lambda=0.1$ were used in combination with
10 substitute model training epochs after each augmentation step. The oracle was
again trained for 15 epochs for MNIST, and 20 epochs for CIFAR-10.

%\vspace{-5mm}

\begin{table}[h!]
\centering
\iffalse
\caption{Oracle adversarial test accuracy on MNIST for
examples generated using a~\cite{Papernot17} style smooth substitute model
black-box misclassification attack with FGSM and $\epsilon=0.3$. Model
`S+*' was trained with 20 iterations of PGD for 40 epochs.}
\fi
\caption{Accuracy of Oracle models on adversarial MNIST examples transferred
	from a~\cite{Papernot17} style smooth substitute model black-box
	misclassification attack. Images from test set attacked with FGSM with
	$\epsilon=0.3$. FGSM adversarial training indicated by '+' suffix, and 20
	iterations of PGD training for 40 epochs by '+*' suffix.}

\label{bb-mnist}
\begin{tabular}{|l|c|c|c|}
\hline
Filters & 64         & 128        & 256         \\ \hline
A       & 79$\pm$1\% & 78$\pm$4\% & 73$\pm$5\%  \\ \hline
A+      & 73$\pm$2\% & 76$\pm$4\% & 80$\pm$2\%  \\ \hline
A+*     & \textbf{95.8$\pm$0.4\%} & \textbf{96.4$\pm$0.3\%} & \textbf{96.7$\pm$0.3\%}  \\ \hline
B       & 46$\pm$5\% & 55$\pm$4\% & 39$\pm$3\%  \\ \hline
B+      & 42$\pm$2\% & 52$\pm$3\% & 50$\pm$6\%  \\ \hline
C       & 51$\pm$4\% & 56$\pm$6\% & 54$\pm$10\% \\ \hline
C+      & 65$\pm$9\% & 72$\pm$6\% & 70$\pm$2\%  \\ \hline
C+*     & \textbf{94.7$\pm$0.3\%} & \textbf{95.6$\pm$0.1\%} & \textbf{96.4$\pm$0.4\%}  \\ \hline
S+*     & 56$\pm$1\% & 68$\pm$2\% & 77.9$\pm$0.7\%  \\ \hline
\end{tabular}
\end{table}

%\vspace{-5mm}

Results for the black-box experiment on the MNIST dataset are shown in
Table~\ref{bb-mnist}. Full-precision networks had a moderate advantage over
undefended binary models B and C. Only the highest capacity full-precision model
benefited from FGSM adversarial training, while the scaled binary model
benefited regardless of capacity. There was a small positive relationship between
accuracy and capacity for both A and C when trained with PGD, and there was 
almost no loss in accuracy in this setting after binarization.
PGD was more effective than stochasticity here as it leads to learning a 
more optimal decision boundary, rather than confusing an adversary with 
dynamic gradient information.

% \usepackage{multirow}
\begin{table}[h!]
\centering
\caption{Accuracy of Oracle models on clean and adversarial CIFAR-10
examples transferred from a~\cite{Papernot17} style smooth substitute model
black-box misclassification attack with FGSM and $\epsilon=0.3$. FGSM
 adversarial training indicated by '+' suffix.}
\label{bb-cifar10}
\begin{tabular}{|c|l|c|c|c|}
\hline
Filters             & Accuracy Type & 64             & 128            & 256            \\ \hline
\multirow{2}{*}{A}  & Clean         & 64.4$\pm$0.6\% & 64.2$\pm$0.3\% & 63.2$\pm$0.9\% \\ \cline{2-5}
                    & Transfer      & 23$\pm$2\%     & 22$\pm$1\%     & 22$\pm$1\%     \\ \hline
\multirow{2}{*}{A+} & Clean         & 64.1$\pm$0.6\% & 64$\pm$1\%     & 65.2$\pm$0.4\% \\ \cline{2-5}
                    & Transfer      & 16.8$\pm$0.7\% & 22$\pm$1\%     & 19$\pm$1\%     \\ \hline
\multirow{2}{*}{C}  & Clean         & 62$\pm$1\%     & 64$\pm$1\%     & 61$\pm$1\%     \\ \cline{2-5}
                    & Transfer      & 20.2$\pm$0.6\% & 20$\pm$1\%     & 21$\pm$1\%     \\ \hline
\multirow{2}{*}{C+} & Clean         & 57.3$\pm$0.2\% & 61$\pm$1\%     & 63$\pm$2\%     \\ \cline{2-5}
                    & Transfer      & \textbf{24.1$\pm$0.5\%} &
                    \textbf{25.8$\pm$0.5\%} & \textbf{27.6$\pm$0.9\%} \\ \hline
\end{tabular}
\end{table}

%\vspace{-1mm}

\section{Discussion}

%\vspace{-1mm}

We suspect that plain BNNs implement two different kinds of gradient masking.
We discovered the first by tracking the L1 norm of the hidden layer activations
and unscaled logits. BNNs operate with larger range and variance
than `normal' networks, which can be explained by virtue of convolving inputs
with greater magnitude ($\pm1$) compared with the typically small values taken 
by weights and activations. For our 64
kernel CNN, the logits were about 4$\times$ larger than the scaled or 
full-precision networks. This is analogous to the more complex defensive
distillation procedure in which the model to be secured is trained with
soft-labels generated by a \emph{teacher} model. When training the
teacher, a softmax temperature, $T$ $\gg$ 1 is used.
The distilled model is trained on the labels assigned by the teacher and using
the same $T$. At test time, the model is deployed with $T=1$, which causes the
logits to explode with respect to their learned values. The logits
saturate the softmax function and cause gradients to vanish, leading FGSM and
JSMA to fail at a higher rate. However, this defense is defeated with a 
close enough guess for $T$, or via a black box attack~\citep{carlini2017towards}.

The second type of gradient masking is less easily overcome, and has to do with
gradients being inherently discontinuous and non-smooth, as seen in
Figure~\ref{fig:mlp-2x2} of Appendix~\ref{sec:mlp-toy-and}. We believe
that this effect is what gives scaled BNNs an advantage over full-precision
with respect to targeted attacks. Even more importantly, through a 
regularization effect, the decision boundary for the MLP with binary units
(Figure~\ref{fig:mlp-2x2}) better represents the actual function to be 
learned, and is less susceptible to adversarial examples. 
But why does gradient masking have a disproportionate effect when attacking 
compared with training on clean inputs? Models `A' and
`B' were trained to within 1.2\% test accuracy, while `B' had improvements of
9.0\% and 29.5\% on JSMA and CWL2 attacks respectively, corresponding
to $8\times$ and $25\times$ difference in accuracy, respectively, for
adversarial vs.~clean inputs. For JSMA, the performance gap can be 
attributed to the sub-optimality of the attack as it uses logits rather 
than softmax probabilities. Furthermore, to achieve its L0 goal, pairs of 
individual pixels are manipulated which is noisy process in a binarized model.

The success of model `S' with stochastically quantized weights in its third
convolutional layer against iterative attacks is more easily explained.
Adversarial examples are not random noise, and do not occur in random
directions. In fact, neural networks are extremely robust to large amounts of
benign noise. An iterative attack that attempts to fool our stochastically
quantized model faces a unique model at every step, with unique gradients.
Thus, the direction that minimizes the probability of the true class in the
first iteration is unlikely to be the same in the second. An iterative attack
making $n$ steps is essentially attacking an ensemble of $n$ models. By making
a series of small random steps, the adversary is sent on the equivalent of a
wild goose chase and has a difficult time making progress in any particularly
relevant direction to cause an adversarial example.

%\vspace{-2mm}

\section{Conclusion}

%\vspace{-1mm}

We have shown that for binarized neural networks, difficulty in training leads
to difficulty when attacking. Although we did not observe a substantial
improvement in robustness to \emph{single step} attacks through binarization,
by introducing stochasticity we have reduced the impact of the strongest
attacks. Stochastic quantization is clearly far more computationally and memory
efficient than a traditional ensemble of neural networks, and could be run
entirely on a micro-controller with a pseudo random number generator. Our
adversarial accuracy on MNIST against the best white-box attack
(CWL2) is 71$\pm$2\% (S64+) compared with the best full-precision model
1.8$\pm$0.9\% (A256+). Black-box results were competitive between binary and
full-precision on MNIST, and binary models were slightly more robust for
CIFAR-10, which we attribute to their improved regularization. Beyond their 
favourable speed and resource usage, we have demonstrated another benefit of 
deploying binary neural networks in industrial settings. Future work will 
consider other types of low-precision models as well as other adversarial 
attack methods.

\iffinal

\subsubsection*{Acknowledgments}

The authors wish to acknowledge the financial support of NSERC, CFI and
 CIFAR\@. The authors also acknowledge hardware support from NVIDIA and Compute 
 Canada. We thank Brittany Reiche for helpful edits and suggestions that 
 improved the clarity of our manuscript.

\fi

%Use unnumbered third level headings for the acknowledgments. All
%acknowledgments, including those to funding agencies, go at the end of the
%paper.

\clearpage

\bibliography{iclr2018_conference}
\bibliographystyle{iclr2018_conference}

\clearpage

\appendix
\section{Clean Test Error Rates}
\label{sec:clean-test-error}

% note, although the variable number of decimal places in the following table might not be
% aesthetically pleasing, they respect the actual measurement uncertainty. There was
% less uncertainty in training float models than binary. You generally only
% want 1 significant digit in the uncertainty measure.
% http://web.ics.purdue.edu/~lewicki/physics218/significant
\begin{table}[h!]
\centering
\begin{tabular}{|c|c|c|c|}
\hline
\backslashbox{Model}{$K_{Layer1}$} & 64 & 128 & 256    \\ \hline
A       & 1.2$\pm$0.2\%   & 1.1$\pm$0.1\%   & 1.06$\pm$0.2\%  \\ \hline
A+      & 0.99$\pm$0.02\% & 1.0$\pm$0.1\%   & 1.03$\pm$0.03\% \\ \hline
B       & 2.3$\pm$0.1\%   & 2.2$\pm$0.2\%   & 2.3$\pm$0.2\%   \\ \hline
B+      & 2.0$\pm$0.2\%   & 1.73$\pm$0.09\% & 1.9$\pm$0.1\%   \\ \hline
C       & 1.3$\pm$0.2\%   & 1.2$\pm$0.1\%   & 1.2$\pm$0.1\%   \\ \hline
C+      & 1.3$\pm$0.1\%   & 1.21$\pm$0.09\% & 1.08$\pm$0.05\% \\ \hline
\end{tabular}
\caption{Error on clean MNIST test set for models with varying
	capacity and precision. A is full-precision, B is binary, and C is binary
	with a learned scalar applied to the ReLU in hidden layers. All models were
	trained with Adam for 15 epochs with a batch size of 128 and a learning
	rate of 1e-3. For adversarially trained models, we used 20 iterations of
	PGD~\citep{madryetal} with $\epsilon = 0.3$ for the last 5 epochs.}
\label{table:models-and-clean-acc}
\end{table}

\section{MLP Toy AND Problem}
\label{sec:mlp-toy-and}

\begin{figure}[h]
\centering{\subfigure[]{\includegraphics[width=2.5in]{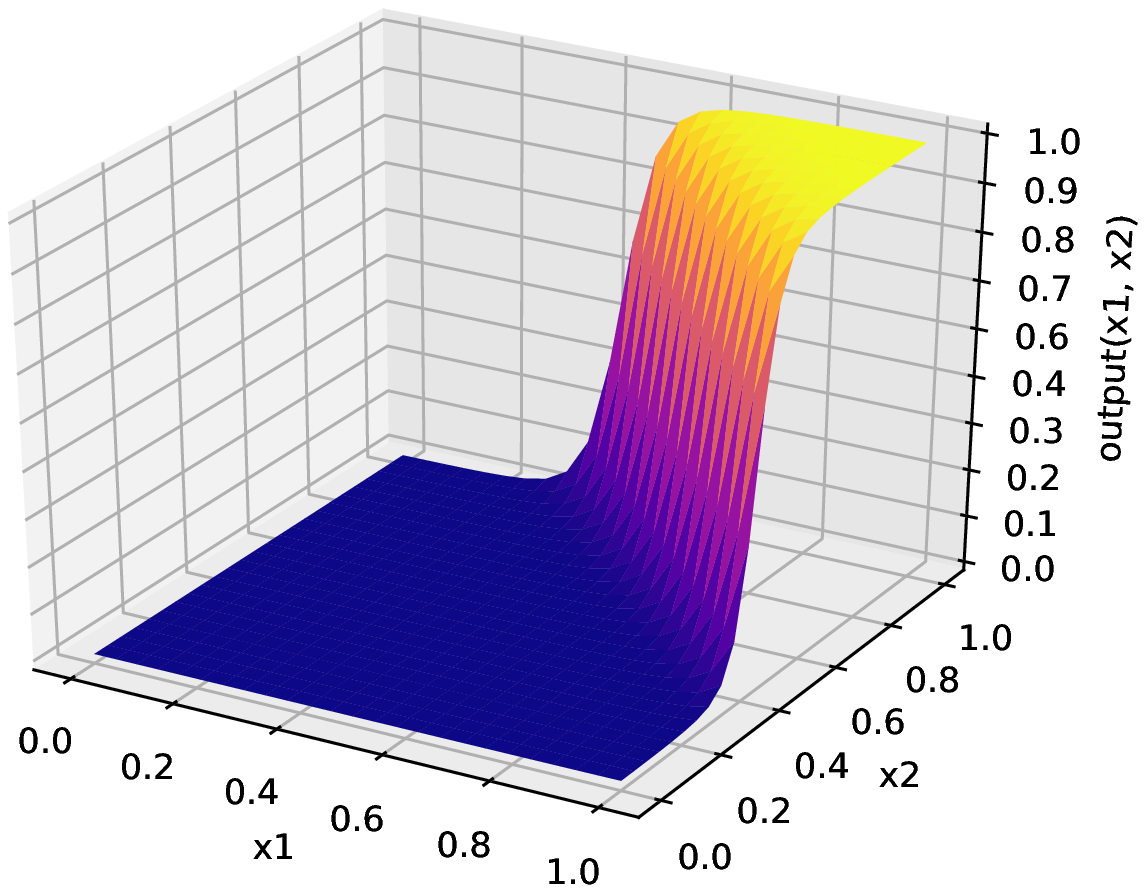}\label{fig:mlp-a}}}
\centering{\subfigure[]{\includegraphics[width=2.5in]{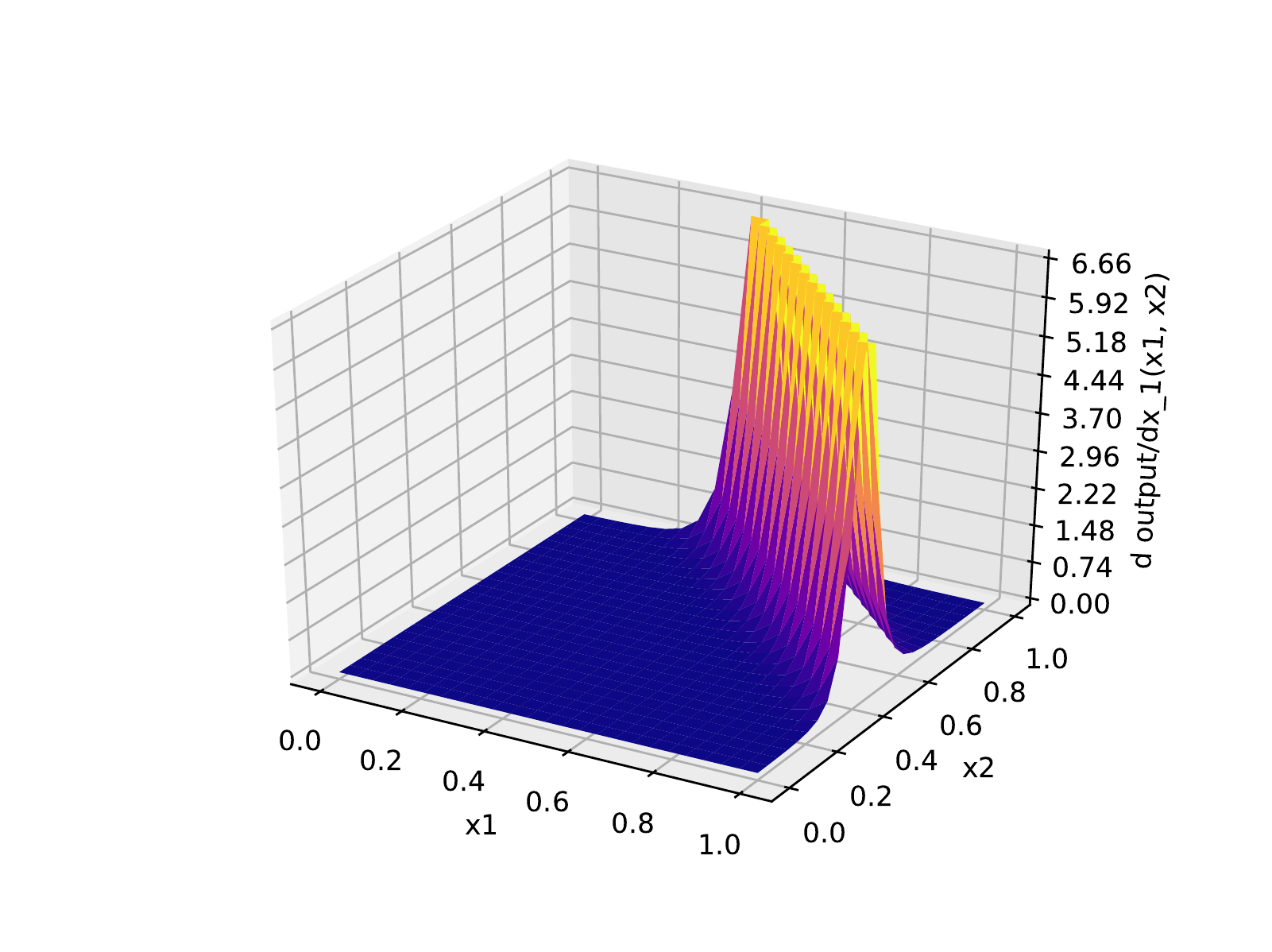}\label{fig:mlp-b}}}
\centering{\subfigure[]{\includegraphics[width=2.5in]{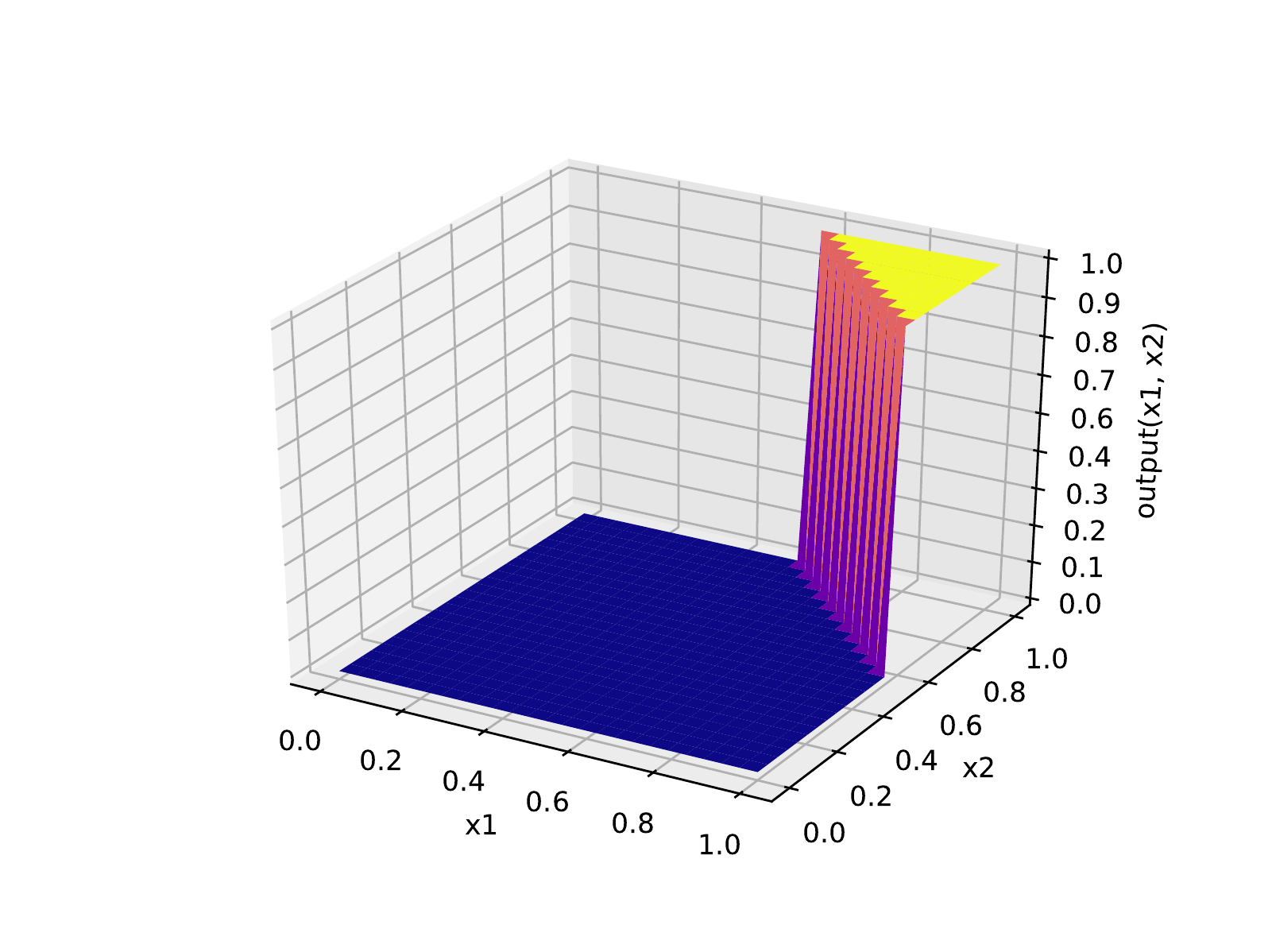}\label{fig:mlp-c}}}
\centering{\subfigure[]{\includegraphics[width=2.5in]{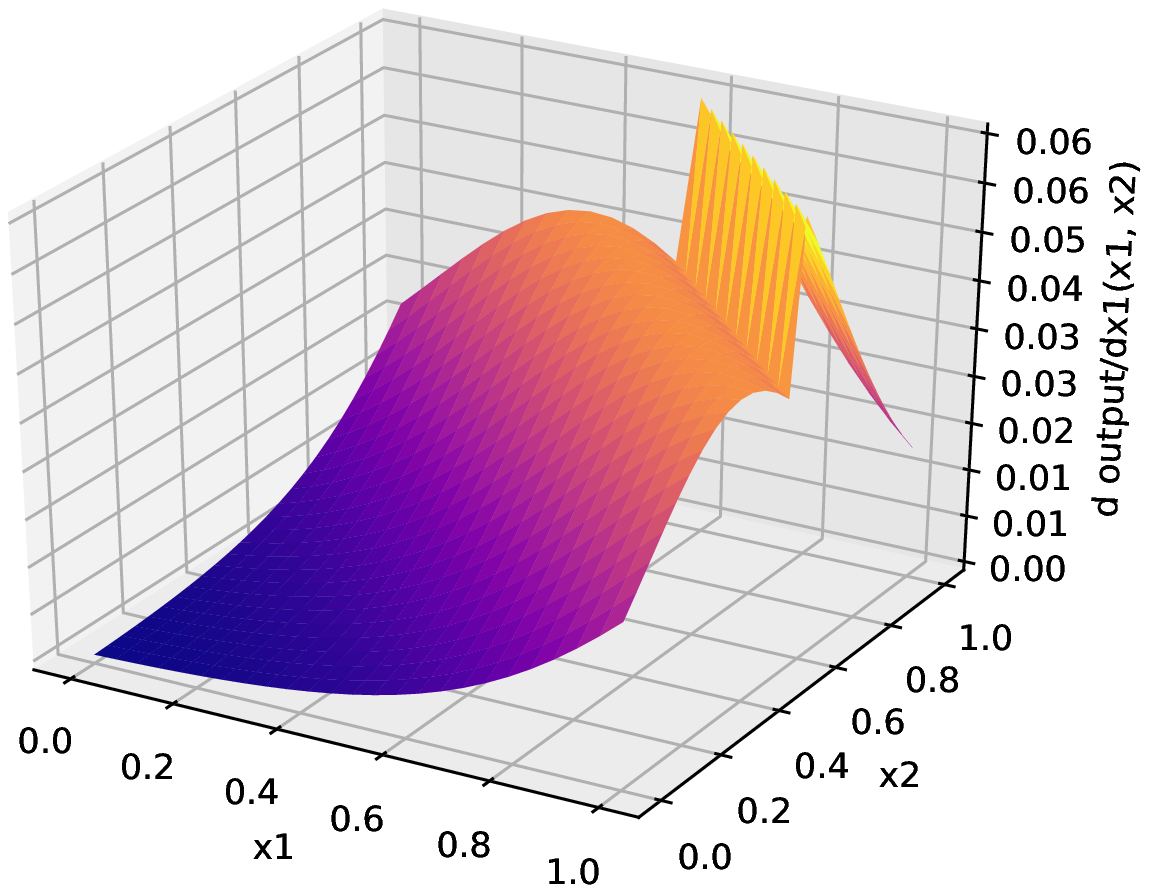}\label{fig:mlp-d}}}
\caption{Decision surface for a three layer MLP with two hidden
units in first two layers, and sigmoid output neuron [\subref{fig:mlp-a} and \subref{fig:mlp-c}]. Corresponding forward derivative with respect to input $x_2$
[\subref{fig:mlp-b} and \subref{fig:mlp-d}]. Full-precision model [\subref{fig:mlp-a} and \subref{fig:mlp-b}] and model with a binarized hidden layer [\subref{fig:mlp-c} and \subref{fig:mlp-d}].}
\label{fig:mlp-2x2}
\end{figure}

We reproduce the toy problem in~\citet{PapernotMJFCS15} of learning the
two-input logical AND function with a simple MLP having two neurons in each
layer. The only difference between our experiment and the original is that we
train a 3-hidden-layer MLP (as opposed to 2-layers) with the Adam optimizer for 1k
epochs, with a learning rate of 0.1. We use 3 layers since this is the smallest
number of layers where the middle one can be quantized without directly
touching the input or output, which would adversely impact learning.
Here, a ``quantized'' layer means that its weights and activations are
thresholded to +1 and -1, and a straight through estimator~\citep{BengioLC13}
is used to backpropagate gradients for learning.

\begin{figure}[h]
\begin{center}
	\subfigure[]{\includegraphics[trim=50 50 900 50, clip, width=2.5in]{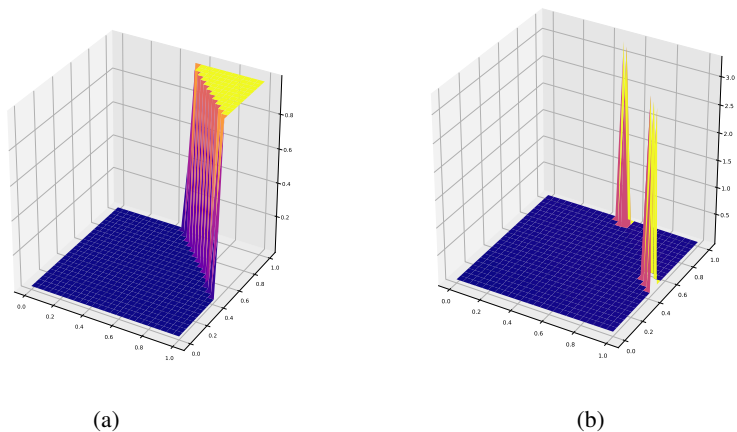}}
	\subfigure[]{\includegraphics[trim=950 50 50 50, clip, width=2.5in]{figures/mlp_binary_and_function_decision_boundary_and_jacobian_1000epochs_4_hidden_1.eps}}
\end{center}
\caption{(a) Decision surface for a 3 layer MLP with four hidden units in first two layers, one output neuron, and quantized middle layer. (b) Corresponding forward derivative.}
\label{fig:mlp-binary-4-hidden}
\end{figure}

All configurations in the AND experiment learn a reasonable decision boundary;
however, the MLPs with a single quantized hidden layer had highly non-linear
forward gradients, as can be seen in Figure~\ref{fig:mlp-d}. As training progresses, the forward derivative was
highly dynamic and took on a variety of different shapes with sharp edges and
peaks. When the MLP was allowed more capacity by doubling the number of hidden
units (see Figure~\ref{fig:mlp-binary-4-hidden}), the forward
derivative was almost entirely destroyed. If one was to use this information to
construct a saliency map, only two regions would be proposed (with poor
directional information), and once exhausted there would be no further choices
more insightful than random guessing.

\clearpage

\section{Visualizing Logits with Scaling Factors}
\label{sec:viz-logits}

\begin{figure}[h]
\centering{\subfigure[]{\includegraphics[width=2.5in]{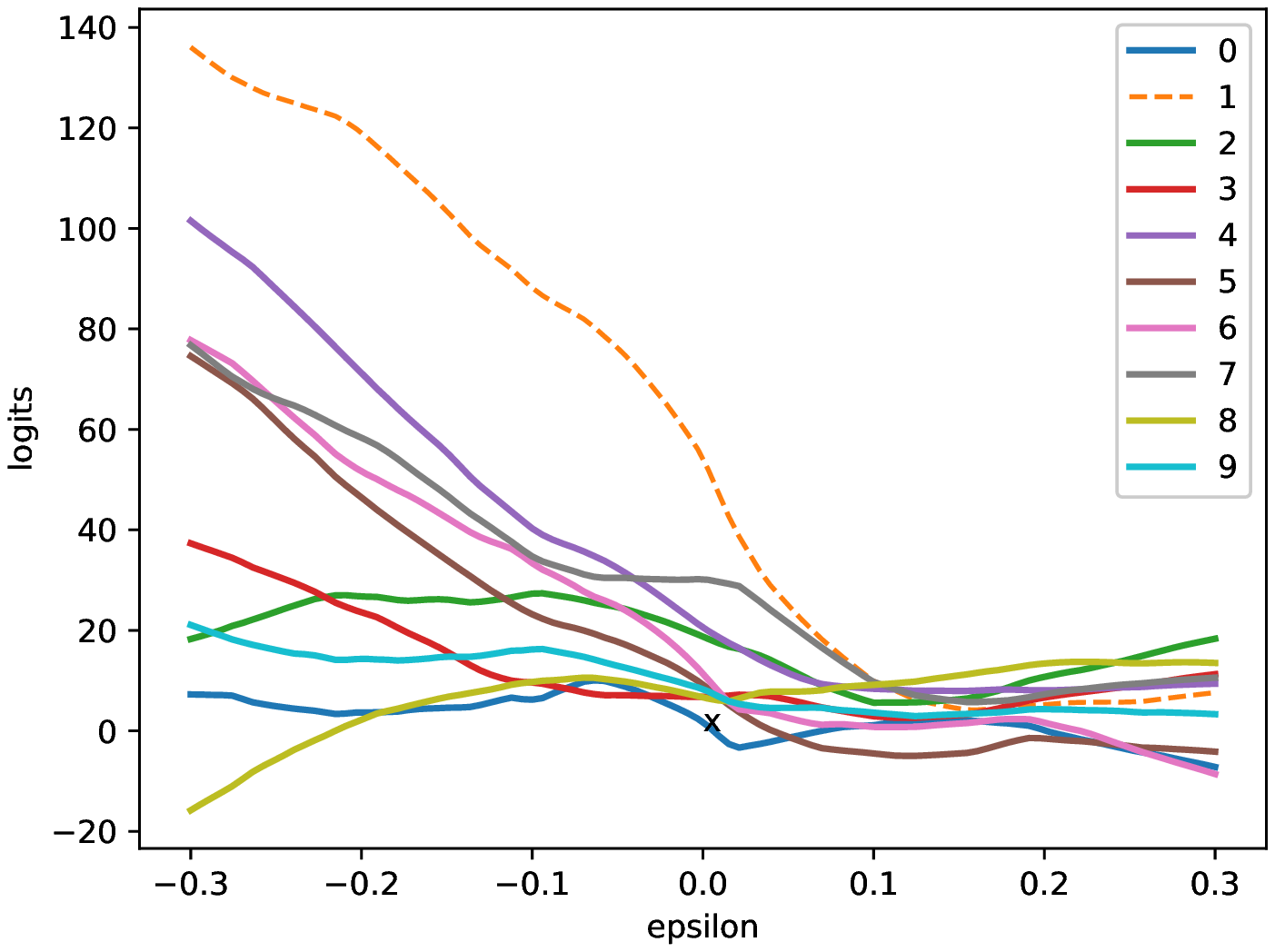}\label{fig:bin_t_06}}} % chktex 8
\centering{\subfigure[]{\includegraphics[width=2.5in]{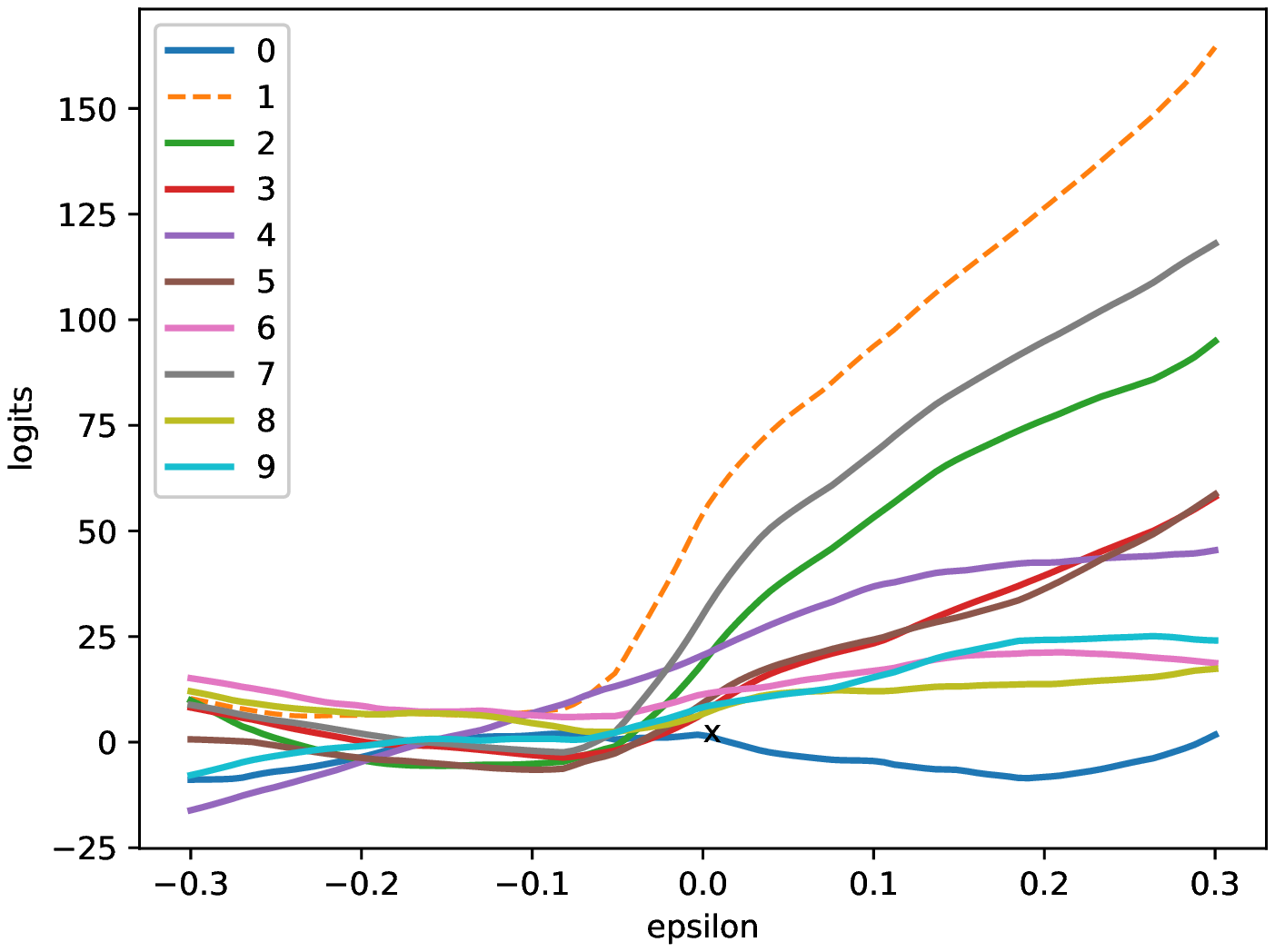}\label{fig:bin_t_07}}} % chktex 8
\centering{\subfigure[]{\includegraphics[width=2.5in]{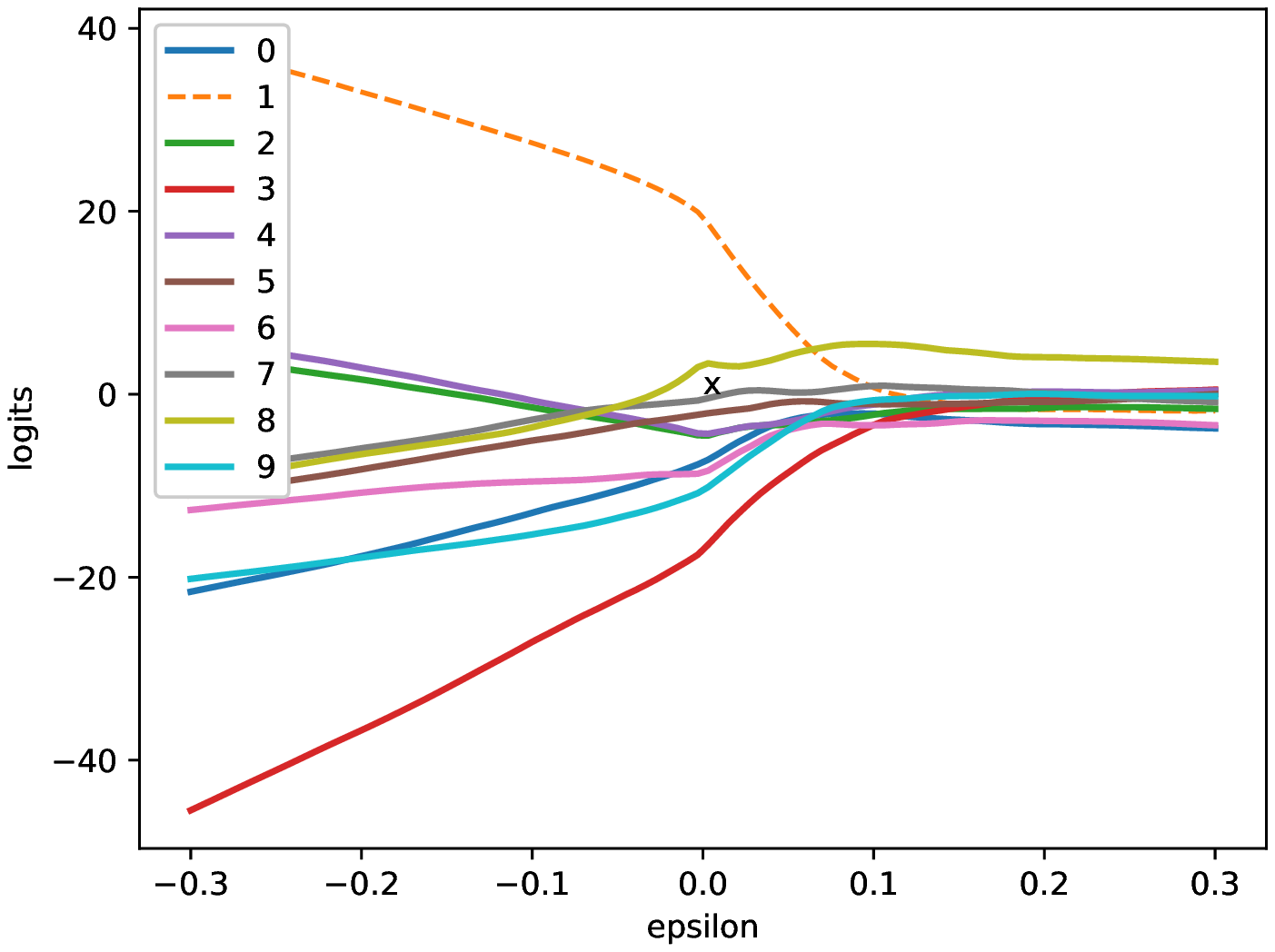}\label{fig:fp_t_01}}} % chktex 8
\centering{\subfigure[]{\includegraphics[width=2.5in]{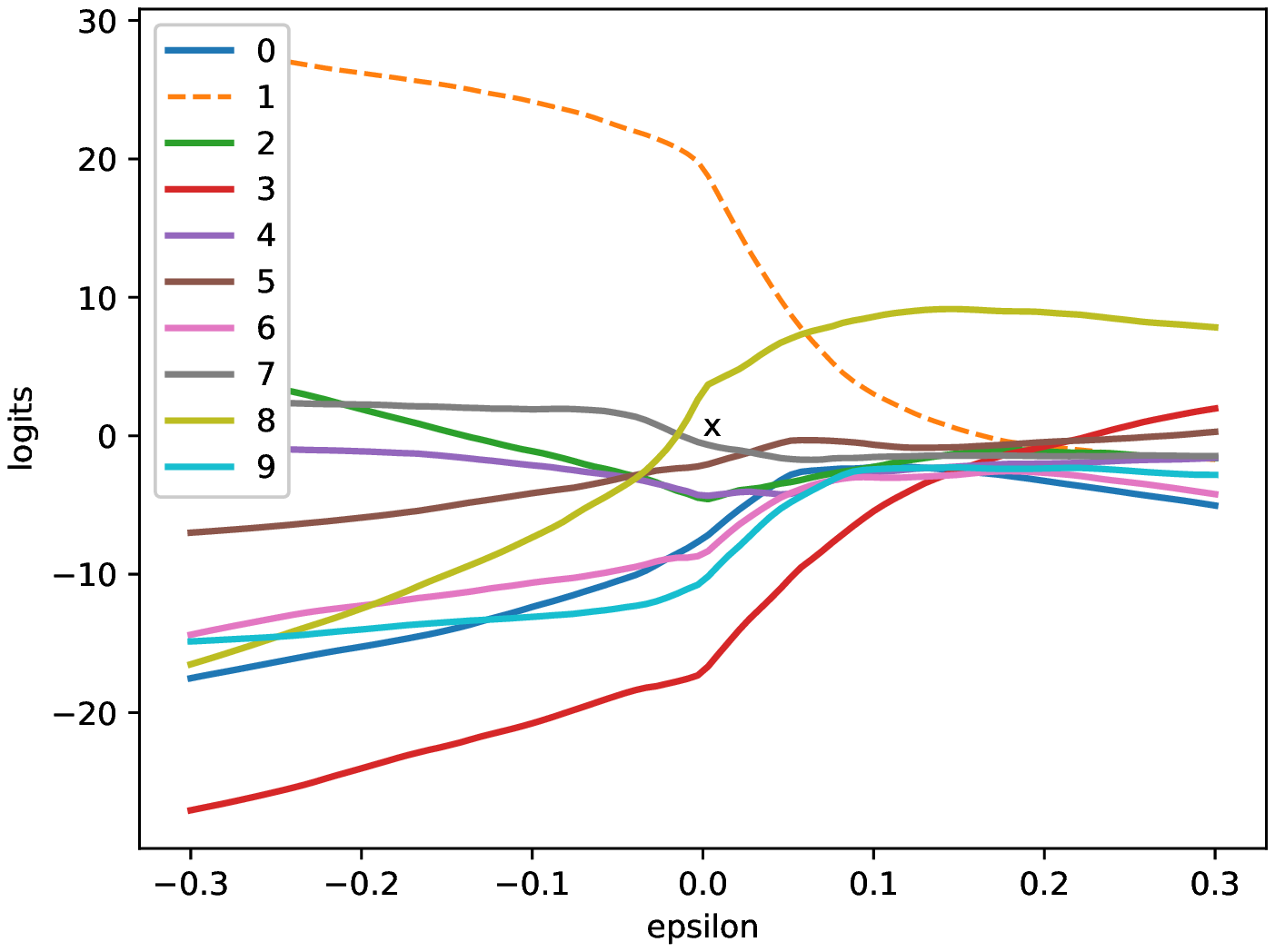}\label{fig:fp_t_10}}} % chktex 8
\caption{We reproduce the plot
	from~\citep{Goodfellow_explaining_adversarial} by evaluating the logits of
	non-scaled binary [\subref{fig:bin_t_06} and \subref{fig:bin_t_07}] and
	full-precision [\subref{fig:fp_t_01} and \subref{fig:fp_t_10}] neural
	networks for an MNIST digit with varying degrees of FGSM perturbation. Note
	that the true class of the digit is ``1'' in this instance. The softmax
	temperature, $T$, was 0.6, 0.7, 0.1, and 1.0 in each of
	\subref{fig:bin_t_06}, \subref{fig:bin_t_07}, \subref{fig:fp_t_01}, and
	\subref{fig:fp_t_10} respectively.}
\label{fig:logits}
\end{figure}

In Figure~\ref{fig:logits} we compare the logits of full-precision and binary
networks under varying degrees of FGSM perturbation. We noticed that for
softmax temperature $T$ between 0.6--0.7 the direction in which increasing the
perturbation causes an adversarial example flips. We observe no similar effect
for full-precision models. Additionally the full-precision logits respond to
scaling in an approximately linear manner, whereas there is very little change
in logits for the binary case apart from the 180 degree flip. We used values
of $\epsilon$ in the range of actual attacks conducted in the paper, however
the piecewise linear effect from~\citep{Goodfellow_explaining_adversarial} is
still there for $\epsilon$ with large absolute value.

\end{document}